\documentclass[lettersize,journal]{IEEEtran}
\ifCLASSOPTIONcompsoc
  \usepackage[nocompress]{cite}
\else
  \usepackage{cite}
\fi
\ifCLASSINFOpdf
\else
\fi
\usepackage{amsmath,amsfonts}
\usepackage{algorithmic}
\usepackage{array}
\usepackage{textcomp}
\usepackage{stfloats}
\usepackage{url}
\usepackage{verbatim}
\usepackage{graphicx}
\usepackage{amssymb}
\usepackage{bm}
\usepackage{subfigure}
\usepackage{multirow}
\usepackage{color}
\usepackage{cite}
\usepackage{tabularx}
\usepackage{float}
\usepackage{fdsymbol}
\usepackage[switch]{lineno}
\usepackage{hyperref}
\usepackage{bbding}
\usepackage{booktabs}

\def\P{{X}}
\def\R{\mathbb{R}}
\def\p{{p}}

\def\x{{x}}
\def\z{{z}}

\begin{document}

\title{Asynchronous Feedback Network for Perceptual Point Cloud Quality Assessment}

\markboth{Journal of \LaTeX\ Class Files,~Vol.~14, No.~8, August~2015}%
{Shell \MakeLowercase{\textit{et al.}}: Bare Demo of IEEEtran.cls for Computer Society Journals}
%



\author{Yujie Zhang, Qi Yang, Ziyu Shan, and Yiling Xu,~\IEEEmembership{Member,~IEEE}
\thanks{This paper is supported in part by National Natural Science Foundation of China (62371290, U20A20185) and  the Fundamental Research Funds for the Central Universities of China, and Science and Technology Commission of Shanghai Municipality (STCSM) under Grant (22DZ2229005). The corresponding author is Yiling Xu (e-mail: yl.xu@sjtu.edu.cn). }
\thanks{Y. Zhang, Z. Shan, Y. Xu are from Cooperative Medianet Innovation Center, Shanghai Jiao Tong University, Shanghai, 200240, China, (e-mail: yujie19981026@sjtu.edu.cn, shanziyu@sjtu.edu.cn, yl.xu@sjtu.edu.cn)}
\thanks{Q. Yang is with University of Missouri–Kansas City, Kansa city, America (email: littlleempty@gmail.com) }

}

\IEEEtitleabstractindextext{%
\begin{abstract}
Recent years have witnessed the success of the deep learning-based technique in research of no-reference point cloud quality assessment (NR-PCQA). For a more accurate quality prediction, many previous studies have attempted to capture global and local features in a bottom-up manner, but ignored the interaction and promotion between them. To solve this problem, we propose a novel asynchronous feedback quality prediction network (AFQ-Net). Motivated by human visual perception mechanisms, AFQ-Net employs a dual-branch structure to deal with global and local features, simulating the left and right hemispheres of the human brain, and constructs a feedback module between them. Specifically, the input point clouds are first fed into a transformer-based global encoder to generate the attention maps that highlight these semantically rich regions, followed by being merged into the global feature. Then, we utilize the generated attention maps to perform dynamic convolution for different semantic regions and obtain the local feature. Finally, a coarse-to-fine strategy is adopted to merge the two features into the final quality score. We conduct comprehensive experiments on three datasets and achieve superior performance over the state-of-the-art approaches on all of these datasets. The code will be available at \url{https://github.com/zhangyujie-1998/AFQ-Net}.

\end{abstract}

}

\maketitle

\IEEEdisplaynontitleabstractindextext

%
\IEEEpeerreviewmaketitle

\section{Introduction}\label{sec:intro}

With recent advancements in 3D capturing devices, point clouds have become a prominent media format for representing 3D visual content in various immersive applications. Point clouds consist of a collection of points distributed in 3D space, where each point has spatial coordinates and additional attributes such as RGB values. Before reaching the user-client, point clouds undergo a wide variety of stages, including acquisition, compression, transmission, and rendering. Any stage might cause degradation in visual quality. Accordingly, it is essential to develop an effective metric that pervades human perception into the research of point cloud quality assessment (PCQA), especially in the common no-reference (NR) situation where pristine reference point clouds are not available. 

\begin{figure}
  \centering 

    \subfigure[]{
    \includegraphics[width=0.42\linewidth]{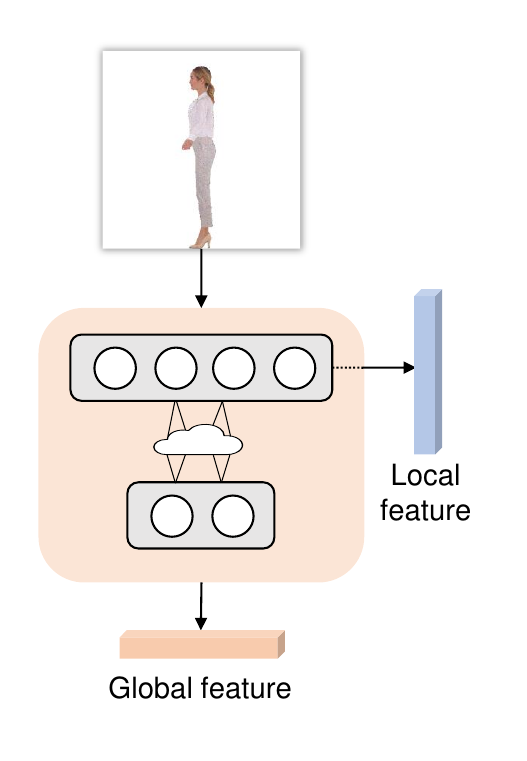}}
    \subfigure[]{
    \includegraphics[width=0.54\linewidth]{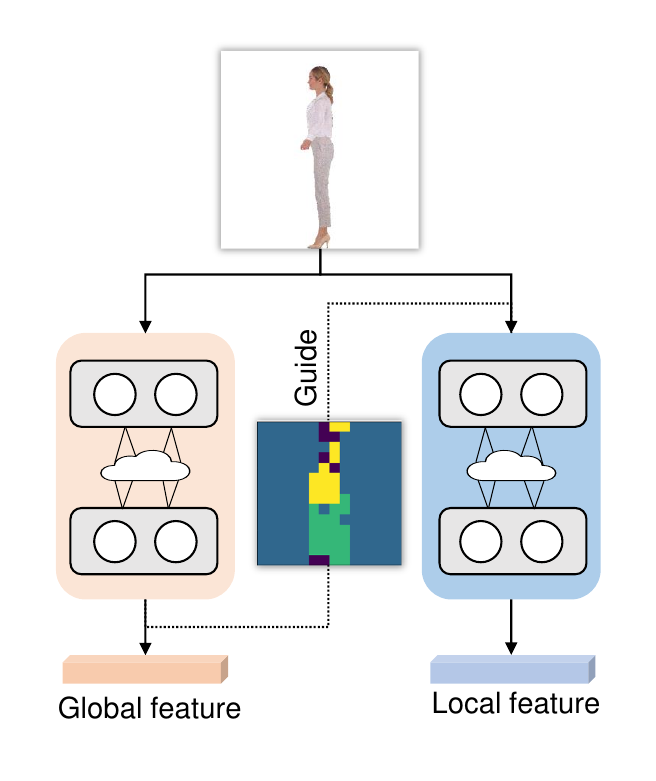}}

  \caption{The motivation of this paper. The core idea is that the human brain can initially identify various semantic regions (\textit{e.g.}, head, trunks, or background) and then perform fine-grained observation for different regions. Therefore, unlike previous studies that extract features in a bottom-up manner, our method applies a dual-branch architecture with a feedback connection. (a) Feature learning paradigm in previous works. (b) Proposed feature learning strategy.}
  \label{fig:motivation}
 
\end{figure}

In the past years, deep-learning-based PCQA methods \cite{liu2021pqa,tliba2023novel,shan2023gpa,Zhang2022MMPCQA, tu2022v, xie2023pmbqa, zhang2023gms, zhu20243dta, mu2023multi, chai2024plain, shan2024contrastive, chen2024dynamic} have shown remarkable performance on multiple benchmarks, which can be applied directly for original 3D data or for projected 2D images. To achieve a more robust prediction, some visual-related methods \cite{liu2023point,Zhang2022MMPCQA,shan2023gpa,xie2023pmbqa, su2023support} advertise learning discriminative features, where two types of feature are usually considered: global feature and local feature. 
Many previous studies \cite{liu2023point,Zhang2022MMPCQA}, whether based on 3D or 2D operations, have sought to extract the two features in a bottom-up manner, seeing Fig. \ref{fig:motivation} (a). Given a multi-layer neural network, features of shallow and deep layers are separately extracted as local and global feature due to the gradually increasing receptive field size. The two types of feature are then simply fused (\textit{e.g.}, concatenation, addition) to regress into the final quality score.


Although promising results have been reported with the existing methods, several problems remain for the utilization of global and local features. First, the bottom-up manner usually depends on a cascaded convolutional structure following a progressive feature extraction order, which fails to exploit the interaction between two types of feature. Actually, quality perception to deformations can vary across different semantical regions, and the sensitivity of a certain local distortion can be reduced or highlighted due to global characteristics \cite{li2018has}. Ignoring the guidance effect of global feature can lead to a limited ability to capture degradation. Second, current models with the projected image from 3D content as input is easily polluted with background information \cite{tsmd-vcip}. When the model uses a 2D CNN-based network as the backbone, the model intends to perform in a filter-sharing manner for both meaningless image backgrounds and rendered point cloud content, which may impair the final performance and is contradictory to subjective observation. In fact, human observers can easily distinguish the background pixels and focus on the point cloud content, which also can assist humans perceive global and local characteristics. Nevertheless, this aspect is seldom taken into account in most existing methods.

To address these problems, we draw two main inspirations from human visual perception mechanisms: i) recent studies \cite{shinohara2008left,kawakami2003asymmetrical} have highlighted the distinct visual processing of global and local feature in different areas of the brain. Specifically, the left and right hemispheres of the brain are specialized in processing global and local information, respectively. These findings suggest the presence of a dual-stream processing mechanism in our brain, challenging existing methods that seek to capture distinct features using a single branch. ii) Mounting evidence \cite{navon1977forest,cao2015look,zhang2021seeing} has indicated that there exists an \textit{asynchronous} process in visual perception. In general, human visual cortex can first perceive the global feature, which is then used to guide the extraction of the local feature through a feedback connection. For example, when presented with a human body point cloud or its projected images, individuals can initially identify various semantic regions (\textit{e.g.}, head, trunk, or background) and then perform fine-grained observation for different regions. Based on these insights, we expect to design a new feature learning strategy for NR-PCQA to overcome the shortcomings of previous studies.

In this paper, we propose a new NR-PCQA method, called  \textbf{A}synchronous \textbf{F}eedback \textbf{Q}uality Prediction \textbf{Net}work (AFQ-Net). Motivated by the human visual perception mechanisms, AFQ-Net employs a dual-branch structure to deal with global and local feature (called the global branch and the local branch), and a feedback module is constructed between branches, seeing Fig. \ref{fig:motivation} (b). More specifically, the input distorted point cloud is first projected into three types of multi-view images: texture images, depth images, and occupancy images. For the global branch, the texture and depth images are fed into 
a transformer-based encoder to generate the attention maps that highlight these semantically rich regions. The attention maps are then merged into the global feature through an occupancy-weighted multi-view fusion, which can reduce the influence of useless background information.

After obtaining the global feature, a feedback module is established to promote the feature extraction process of the local branch. Concretely, inspired by the merits of region-aware convolution \cite{chen2021dynamic}, we leverage the generated attention maps to perform dynamic convolution for different semantic regions of the original projected images and obtain the needed local feature.
Finally, a coarse-to-fine strategy is adopted to merge the two features into the quality score. We propose a feature disentangling loss to improve feature discrepancy and a branch-wise quality ranking loss to maintain a progressive quality prediction. The latter loss enables the network to first take the coarse prediction from the global feature and gradually approach the finer-grained prediction by refining the local feature. We summarize the main contributions as follows:

\begin{itemize}
    \item We propose a new NR-PCQA method named AFQ-Net, which is consistent with  human visual perception. Through the feedback module between the two branches, the global feature can effectively guide the extraction of local feature. 
    \item We propose a coarse-to-fine quality learning strategy. This strategy enables the network to effectively fuse the global and local feature and to achieve a progressive quality prediction.
    \item We conduct comprehensive experiments on three datasets (SJTU-PCQA\cite{yang2020predicting}, WPC\cite{liu2022perceptual}, LS-PCQA\cite{liu2023point}), and achieve superior performance over the state-of-the-art approaches on all of these datasets.
\end{itemize}

The remainder of this paper proceeds as follows. Section \ref{sec:related work} presents the related work. The implementation details of the proposed method are presented in Section \ref{sec:method}. Section \ref{sec:experiment} gives the experiment results. Conclusion is drawn in Section \ref{sec:conclusion}.

\section{Related Works}\label{sec:related work}

\textbf{Point Cloud Quality Assessment.} NR-PCQA aims to evaluate the perceptual quality of distorted point clouds without available references. NR metrics can be performed either over the 2D projection of point clouds or directly on the raw 3D data. For the projection-based methods,
Liu \textit{et al.} \cite{liu2021pqa} proposed to project point clouds to multiple planes and leverage distortion classification information as an auxiliary feature to assist in the training of the network. Zhang \textit{et al.}\cite{Zhang2022Video} integrated the point cloud projection into a video, followed by the utilization of video quality assessment methods to evaluate the quality of point clouds.  Xie \textit{et al.} \cite{xie2023pmbqa} first obtained four types of projected images (texture, normal, depth, and roughness) and then developed a graph-based feature fusion module to fuse different features. For the methods performed directly on raw 3D data, Liu \textit{et al.} \cite{liu2023point} first transformed point clouds into voxels and then adopted an end-to-end sparse convolution network to learn the quality representation of point clouds. Shan \textit{et al.} \cite{shan2023gpa} modeled point clouds using graph structures and then inferred perceptual quality using a multi-task graph neural network. Tliba \textit{et al.} \cite{tliba2023novel} first divided point clouds into local patches and then employed a PointNet-based \cite{qi2017Pointnet} architecture to extract features. In addition, some methods have been developed to leverage both point cloud projection and raw 3D data to extract integrated features. For example, Zhang \textit{et al.} \cite{Zhang2022MMPCQA} utilized 2D and 3D encoders to separately extract features from point clouds and the corresponding projections and fused features using the attention mechanism. Wang \textit{et al.} \cite{wang2023applying} proposed a collaborative cross-modal adversarial learning strategy to enhance the fused features extracted from two modalities.

Recently, some research \cite{liu2021reduced,liu2022no,su2023support,lv2024no, duan2024perceptual, long2024perceptual} has aimed to measure point cloud quality using payload information extracted from the compressed bitstream, which benefits real-time and nonintrusive quality monitoring. For example, Liu \textit{et al.} \cite{liu2022no} studied the relationship between the perceptual quality and the texture/geometry quantization parameter in video-based point cloud compression (V-PCC) algorithm and proposed a no-reference bitstream-layer model. In \cite{lv2024no}, Lv \textit{et al.} proposed a new texture distortion model for lossless geometric coding in geometry-based point cloud compression (G-PCC) and integrated it with the position quantization scale (PQS) results for quality prediction.

\begin{figure*}[t]
    \centering
    \includegraphics[width=0.95\linewidth]{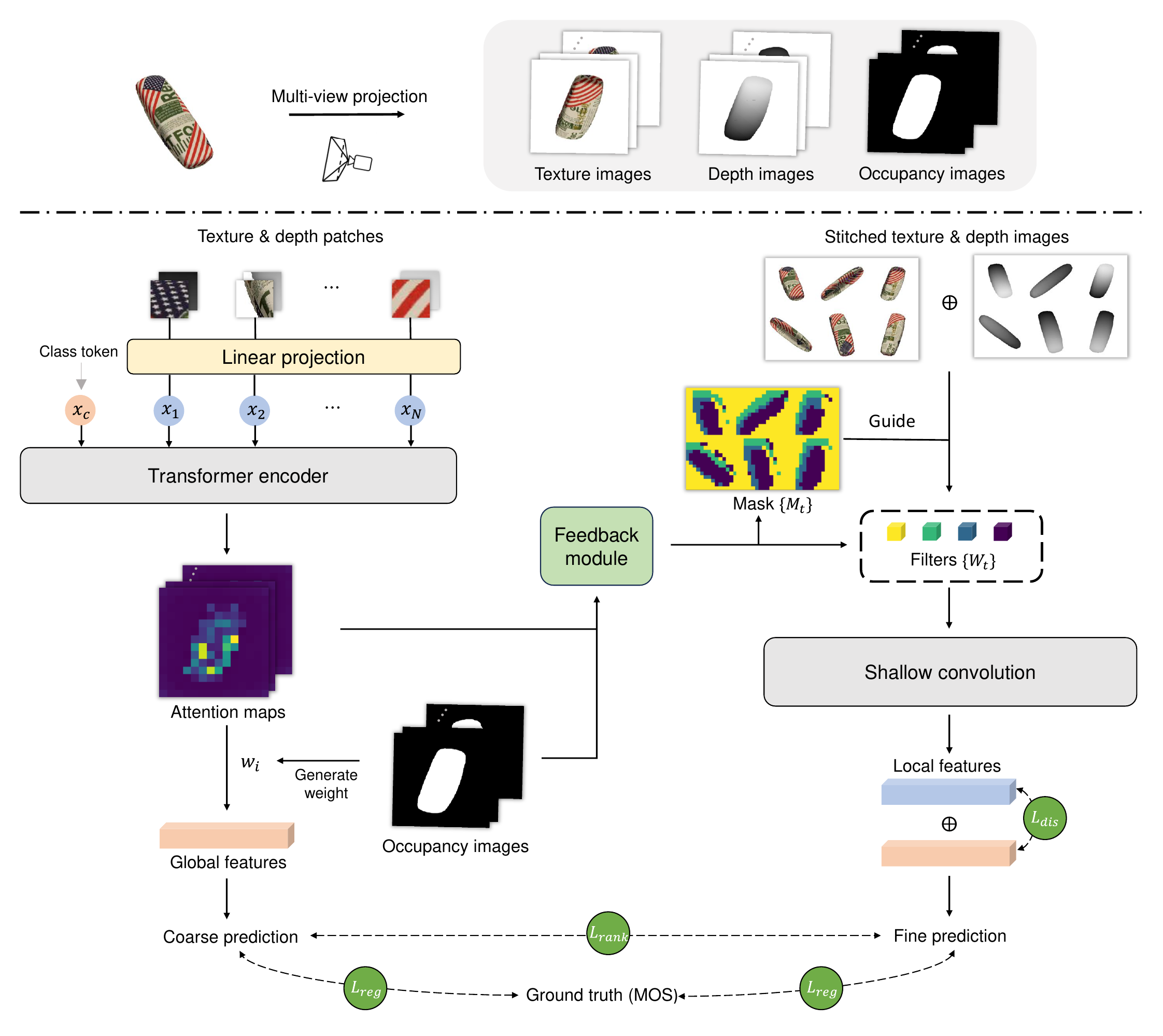}
    \caption{The proposed AFQ-Net framework, which includes two branches and a feedback module, where $\oplus$ denotes the concatenation operation. We first divide images into patches and feed them into a transformer encoder, and the attention maps of class token are merged into the global feature. Then, we utilize the attention maps to generate a guided mask and multiple groups of filter, followed by performing region-aware convolution to derive the local feature. Finally, we employ a coarse-to-fine quality prediction based on the extracted two features.}
    \label{fig:framework}

\end{figure*}
\textbf{Global-Local Feature Learning.} Global-local feature learning is a prevalent strategy in quality assessment research, which is based on the fact that perceptual quality is determined by both high-level semantic and low-level details. Many previous studies \cite{liu2023point,Zhang2022MMPCQA} separately extract the output of shallow and deep
layers of networks as local and global feature and perform parallel processing, ignoring the interaction and promotion between features. In image/video quality assessment scenarios, researchers have proposed several methods to better utilize the two types of feature. An intriguing way \cite{su2020blindly} is to utilize the global feature to generate a hyper network \cite{ha2016hypernetworks} to replace the common quality regression module, where the hyper network represents the perception rule for a specific sample. In \cite{chen2023topiq} and \cite{hu2023blind}, Chen \textit{et al.} and Hu \textit{et al.} both proposed cross-scale fusion mechanisms that allow information to propagate between features. In \cite{chen2021learning}, Chen \textit{et al.} proposed an attention module
to assign weight for multi-scale features by their importance. However, the above improvements focus more on feature post-processing, that is, how to increase the interaction between already generated features, ignoring the intrinsic temporal order of feature extraction in brain mechanisms. In comparison, we intend to use global feature directly to guide the ``\textit{generation}" of local feature, which is an asynchronous process and shows more conformity to visual perception.

\section{Method} \label{sec:method}
\subsection{Overview}

The aforementioned analysis reveals that human perception is a dual-stream asynchronous process, and global feature are usually first perceived to guide the extraction of local feature. On the basis of the philosophy, as illustrated in Fig. \ref{fig:framework}, the framework of our model is divided into three stages: i) First, global feature are extracted based on the projected three types of multi-view point cloud images (\textit{i.e.}, texture, depth, and occupancy). We divide the texture and depth images into patches and feed them into one vision transformer (ViT) \cite{dosovitskiy2020image} pre-trained on ImageNet \cite{deng2009imagenet}. The attention maps of the class token (a learnable embedding for semantic prediction) in the ViT are merged into the global feature. ii) Second, we utilize the attention maps to guide the extraction of local feature. We use the attention maps to generate a mask map and several regional filters, the former highlighting multiple semantic regions. 
Based on the generated mask map and regional filters, we perform dynamic convolution for different regions of the original projected images and obtain the compact local feature, which are then fused with the global feature. iii) Third, we perform a coarse-to-fine quality prediction. The global and fused features are both supervised by the overall quality score, and a branch-wise quality ranking loss is applied to guarantee a progressive quality prediction.

\begin{figure*}[t]
    \centering
    \subfigure[]{\includegraphics[width=0.32\linewidth]{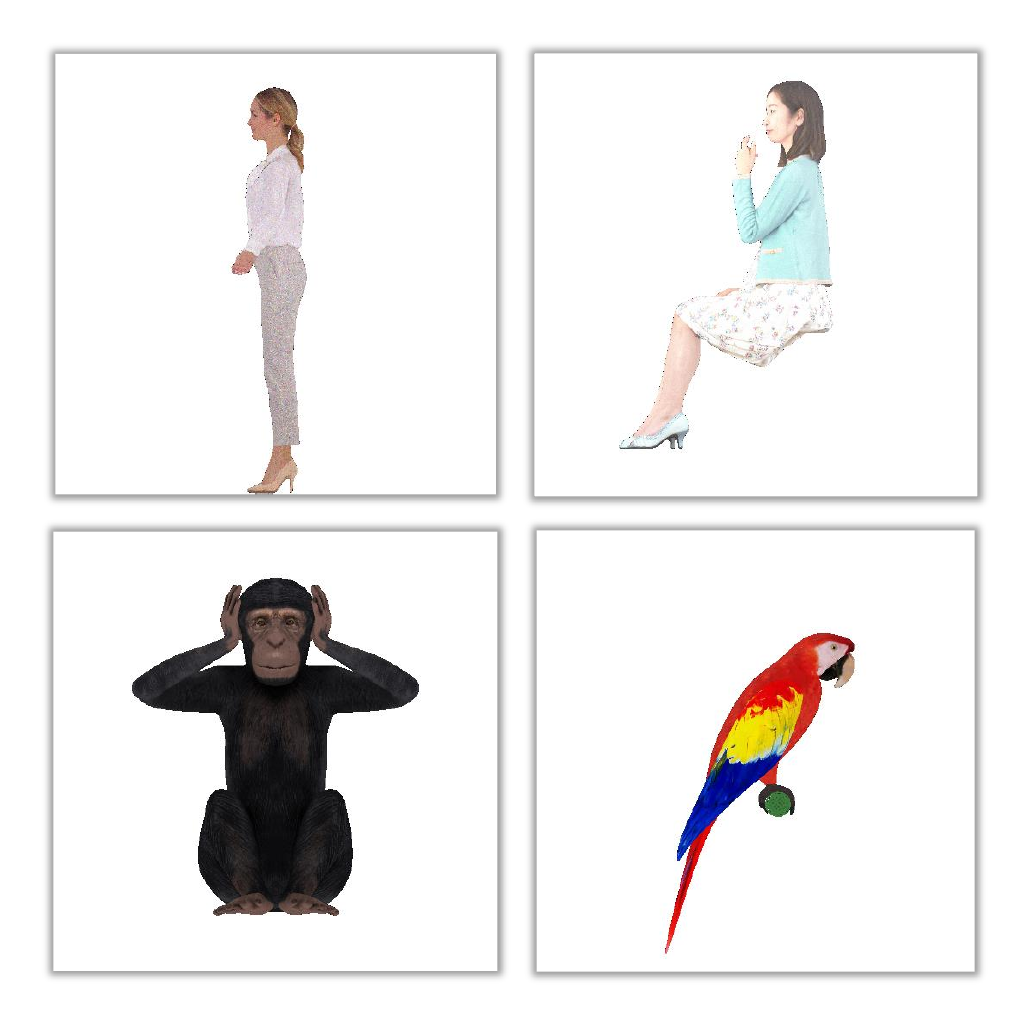}}
    \subfigure[]    {\includegraphics[width=0.32\linewidth]{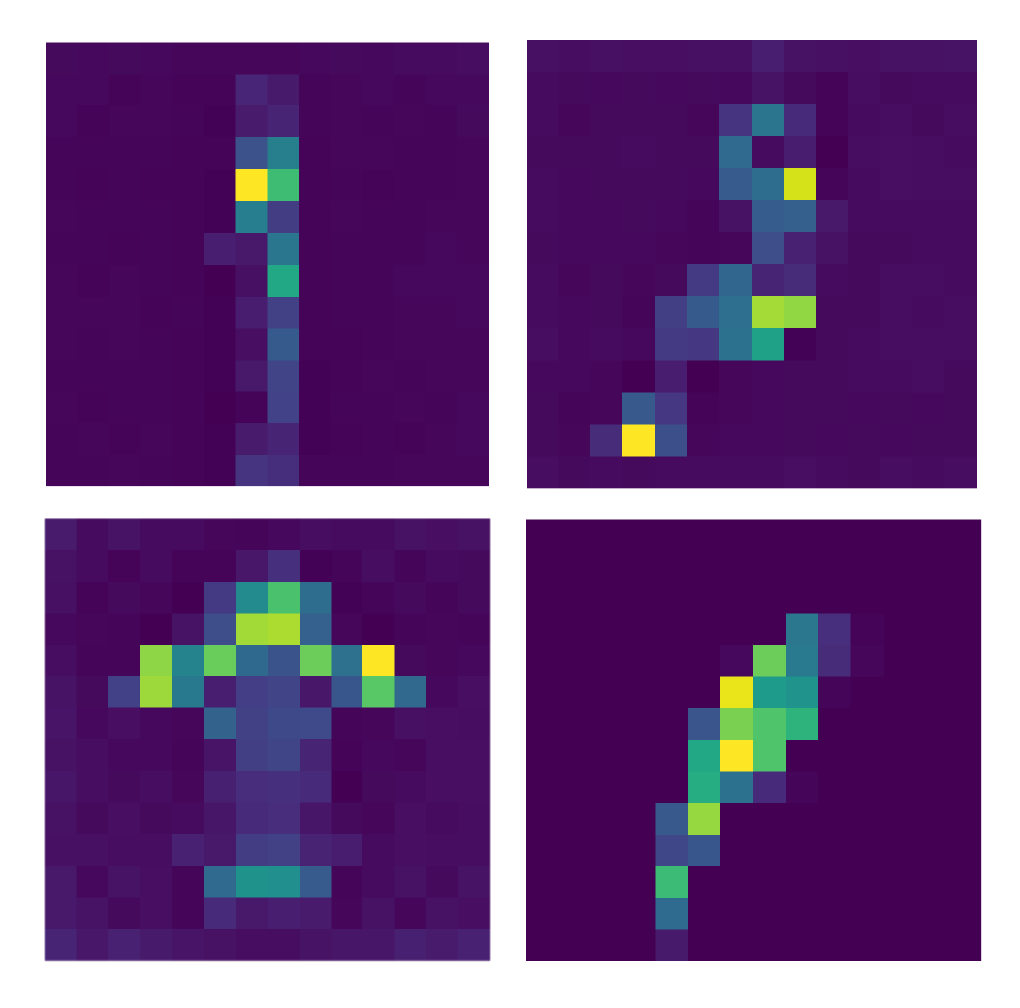}}
    \subfigure[]    {\includegraphics[width=0.32\linewidth]{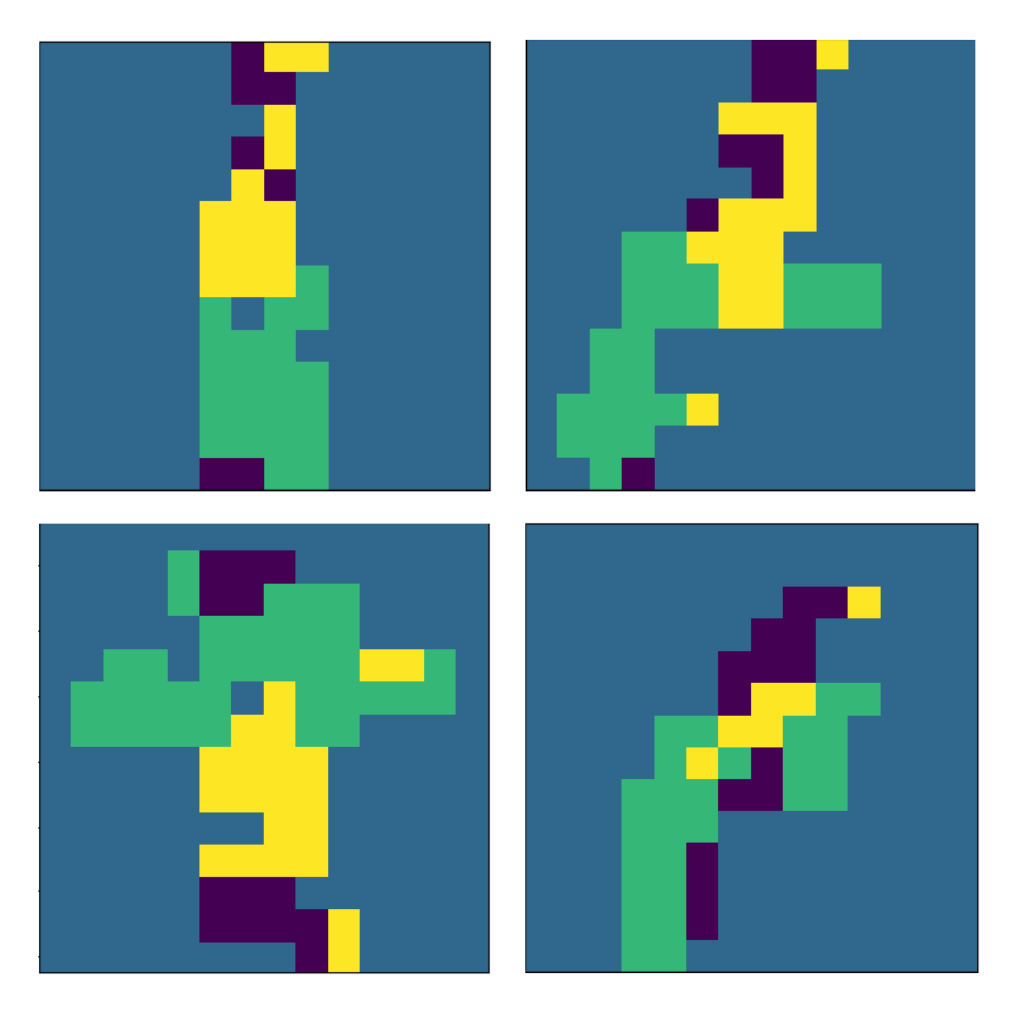}}
    \caption{Visusalization of the attention maps corresponding to the class token and the generated mask. (a) Original projected texture images. (b) Attention maps corresponding to the class token. (c) Guided masks that divide the images into multiple semantic regions. Different colors in (b) indicate the size of the element value in the attention map; the brighter the color, the larger the element value. In (c), colors are used to distinguish different regions, and pixels of the same color are considered to have the same semantics. }
    \label{fig:attention_map}
    
\end{figure*}

\subsection{Preprocessing}
Instead of inferring point cloud quality directly in 3D space, we project point clouds into 2D images to better leverage mature 2D pre-trained models targeted for semantic-related tasks (\textit{e.g.}, classification). 
Given a point cloud $X$, we render it into multi-view images from six perpendicular perspectives (that is, along the positive and negative directions of the x,y,z axes). To reduce information loss during the projection process, we obtain three different types of image to represent $X$: texture images, depth images, and occupancy images. More specifically, we use the \texttt{PointsRenderer} function in the Pytorch3D \cite{ravi2020accelerating} library for point cloud projection. The renderer uses an orthographic camera, and applies alpha compositing to generate the pixel values (please refer to \cite{wiles2020synsin} for more details).
The acquired former two types of image reflect the original color and geometry information and are normalized to the range $[0,1]$. The occupancy images are binary images locating the point cloud contents, where ``1" indicates the point cloud contents while ``0" indicates background areas.
We separately denote the three types of multi-view images of $X$ as $\{X_t^i\in \R^{H\times W\times3}|_{i=1}^6\}$, $\{X_d^i\in \R^{H\times W}|_{i=1}^6\}$ and $\{\P_o^i\in \R^{H\times W}|_{i=1}^6\}$, where $i$ denotes the $i$-th viewpoint.


\subsection{Attention-aware Global Feature Extraction}
The vision transformer (ViT) has been shown to be capable of building long-term dependencies and prioritizing low-frequency information more than CNN \cite{ghiasi2022vision,park2022vision}, making it a good choice for global feature extraction. To make the projected images compatible with the ViT, for $\P_t$ and $\P_d$ from an arbitrary perspective, we first partition them into non-overlapping $P\times P$ patches.

\textbf{Encoding by Vision Transformer.}
 Given a texture patch $\p_t$ and its corresponding depth patch $\p_d$, we separately map two patches into two vector representations and simply fuse them with position embedding (denoted by $\mathrm{PE}$), \textit{i.e.},
\begin{equation}
    \x = \mathrm{PatchEmbed}_t(\p_t) + \mathrm{PatchEmbed}_d(\p_d) + \mathrm{PE},
\end{equation}
where $\x\in \R^D$ is the token embedding; $\mathrm{PatchEmbed}_{t(d)}(\cdot)$ are two independent embedding operations.

Next, we follow \cite{dosovitskiy2020image} to prepend a learnable embedding $\x_{c}$ named class token to the sequence of embedded patches. The token sequence can be represented as $\z_0=[\x_{c};\x_1;\x_2;\cdots;\x_N]\in \R^{(N+1)\times D}$, where $N=HW/P^2$ denotes the patch number. We then input $\z_0$ into the ViT, which consists of $L$ consecutive identical transformer blocks and is initialized with the optimized weights on ImageNet. Taking the $l$-th block as an example, it takes the output $\z_{l-1}$ of the previous $(l-1)$-th block as input to compute its output $\z_l$.


\begin{figure}[t]
  \centering 

    \subfigure[]{
    \includegraphics[width=0.9\linewidth]{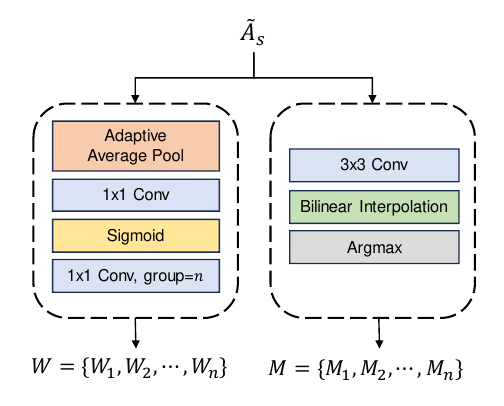}}
    \subfigure[]{
    \includegraphics[width=0.7\linewidth]{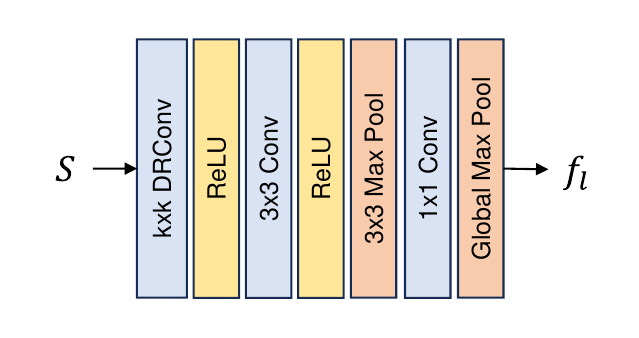}}
  \caption{The specific architecture of the feedback module and the local branch. (a) Feedback module. (b) Local branch.}
  \label{fig:drconv}
\vspace{-0.3cm}
\end{figure}

\textbf{Attention Map Extraction.}
The ViT effectively models the relationship between patches using multi-head self-attention operation $\mathrm{MSA}(\cdot)$ \cite{vaswani2017attention}. For the $j$-th head of $\mathrm{MSA}(\cdot)$, the computational process can formulated as follows:
\begin{equation}
\begin{aligned}
    &[Q_j,K_j,V_j]=\mathrm{LN}(\z_{l-1})\Gamma_j, \Gamma_j\in \R^{D\times 3D_h},\\
    &A_j = \mathrm{softmax}(\frac{Q_jK_j^T}{\sqrt{D_h}}), A_j\in\R^{(N+1)\times(N+1)},\\
    &\mathrm{SA_j}(\z_{l-1}) = A_jV_j ,
\end{aligned}\label{eq:MSA}
\end{equation}
where $\mathrm{SA_j(\cdot)}$ represent the $j$-th self-attention operation, $j\in[1,N_h]$; $\mathrm{LN(\cdot)}$ denote the layer normalization \cite{ba2016layer}; $\Gamma_j$ denotes the projection matrix; $A_j$ denotes the attention matrix. According to Eq. \eqref{eq:MSA}, we can see that each element of the attention matrix $A_j$ indicates the pairwise correlation between two patch representations in $\z_{l-1}$. Furthermore, the class token collects information from all patches via the self-attention operation. In traditional classification tasks, the output of the class token (\textit{i.e.},  $z_{L}[:,1]$) will be transformed into a semantic prediction via a small MLP \cite{dosovitskiy2020image}. Therefore, the correlation between the class token and an arbitrary patch representation can effectively reflect the relevance between the patch and semantics, which facilitates the generation of needed global feature. We define the attention maps $A_c$ of the class token as:
\begin{equation}
\begin{aligned}
    A_c &= \mathrm{LN}(A_c^1\oplus A_c^2\cdots\oplus A_c^{N_h}) \in \R^{\frac{H}{P}\times\frac{W}{P}\times N_h},\\
    A_c^j &= \mathrm{Reshape}(A_j[1,2:N+1])\in \R^{\frac{H}{P}\times\frac{W}{P}},j\in[1,N_h], 
\end{aligned} 
\end{equation}
where $\oplus$ denotes the concatenation operation. To better illustrate the semantic sensitivity of the attention maps, we visualize $A_c$  in Fig. \ref{fig:attention_map} (b), from which we can observe that the attention maps accurately emphasize the content information in the projected images.

\textbf{Global feature Generation.}
After extracting the attention maps of the class token, we further transform $A_c$ in the last transformer block into global feature. Specifically, for the $i$-th viewpoint, the attention maps are processed by a $1\times 1$ convolution and an adaptive average pooling operation to generate a vector representation $f_g^i$. Given the different proportions of content among perspectives, we follow \cite{wu2021subjective} to adopt a simple occupancy-weighted fusion strategy, that is,
\begin{equation}
    f_g = \frac{\sum_{i=1}^6 w_if_g^i}{\sum_{i=1}^6 w_i},
\end{equation}
where $w_i$ denotes the occupancy ratio, that is, the proportion of ``1" in the occupancy image $X_o^i$. We take $f_g\in \R^{D_o}$ as the final global feature.

\begin{table}[]
    \centering
    \caption{\MakeUppercase{THE FEATURE SIZE OF EACH MODULE IN the feedback module and local branch.}}
    \resizebox{0.95\linewidth}{!}{
    \begin{tabular}{c c c c}
    \toprule
         Module & Layer & Input Size & Output Size \\ \midrule
         \multirow{3}{*}[-2.5ex]{Filter Generation} &Avg Pool &$\frac{2H}{P}\times\frac{3W}{P}\times N_h$ &$k\times k\times N_h$  \\ \rule{0pt}{15pt}
         & $1\times1$ Conv &$k\times k\times N_h$ &$k\times k\times n^2 $ \\ \rule{0pt}{15pt}
         & $1\times1$ Conv &$k\times k\times n^2$ &$k\times k\times 64n$ \\ \midrule
         
        \multirow{3}{*}[-2.5ex]{Mask Prediction} &$3\times3$ Conv &$\frac{2H}{P}\times\frac{3W}{P}\times N_h$ &$\frac{2H}{P}\times\frac{3W}{P}\times n$  \\ \rule{0pt}{15pt}
         &Interpolation &$\frac{2H}{P}\times\frac{3W}{P}\times n$ &${2H}\times 3W\times n$ \\ \rule{0pt}{15pt}
         &Argmax &${2H}\times 3W\times n$  &${2H}\times 3W\times 1$  \\ \midrule

        \multirow{3}{*}[-7.5ex]{Local Branch} &$k\times k$ DRConv &$2H\times 3W \times 4$ &$2H\times 3W \times 16$  \\\rule{0pt}{15pt}
         &$3\times3$ Conv &$2H\times 3W \times 16$ &$2H\times 3W \times 64$  \\ \rule{0pt}{15pt}
         &Max Pool &$2H\times 3W \times 64$  &$H\times \frac{3W}{2} \times 64$  \\ \rule{0pt}{15pt}
         &$1\times1$ Conv &$H\times \frac{3W}{2} \times 64$  &$H\times \frac{3W}{2} \times D_o$  \\ \rule{0pt}{15pt}
         &Global Max Pool &$H\times \frac{3W}{2} \times D_o$ &$1\times 1 \times D_o$\\ 
    \bottomrule
    \end{tabular}}
    \label{tab:feature_size}
\end{table}


\subsection{Feedback-guided Local Feature Extraction}

As mentioned in Section \ref{sec:intro}, human observers can easily distinguish different semantic regions with a glance, and feedback connections add details for each region with scrutiny. Therefore, instead of learning only one convolution kernel for all positions, we intend to adaptively generate filters for different regions based on the attention maps. Considering that the same semantic regions may share among different viewpoints, we first concatenate the texture and depth images of $X$ and compose the multi-view images into a stitched image denoted as $S\in\R^{2H\times3W\times4}$, which corresponds to a stitched occupancy image denoted as $S_o\in\R^{2H\times3W}$.

\textbf{Region-aware Feedback.}
In order to dynamically process different semantic regions, we choose DRConv \cite{chen2021dynamic} to take advantage of its idea of producing region-wise filters. Unlike the original DRConv that learns different filters and region divisions directly from the input images, we learn the two terms from the attention maps. This type of design can better exploit the guiding role of global feature. 

Specifically, the attention maps of different viewpoints are first stitched as $A_s\in\R^{\frac{2H}{P}\times\frac{3W}{P}\times N_h}$. Considering that the occupancy images directly distinguish the blank backgrounds using ``0", it is reasonable to merge the occupancy information into the attention maps to achieve better region division. For this purpose, the stitched occupancy image $S_o$ is first resized to the size of $\frac{2H}{P}\times\frac{3W}{P}$, and then multiply with each channel of $A_s$, \textit{i.e.},

\begin{equation}
    \tilde{A}_s = A_s \odot \mathrm{Resize}(S_o)\in \R^{\frac{2H}{P}\times\frac{3W}{P}\times N_h},
\end{equation}
where $\odot$ is element-wise multiplication. 

After enhancing the attention maps with occupancy information, we feed $\tilde{A}_s$ into a dual-stream embedder, which includes two modules: mask prediction module and filter generation module (see Fig. \ref{fig:drconv} (a) for the explicit structure). The former module outputs a guided mask $M=\{M_1,\cdots,M_{n}\}\in\R^{2H\times3W}$ representing the region division, in which pixel values vary from 0 to $n-1$ and indicate $n$ different semantic regions. The latter module generates $n$ groups of $k\times k$ filter as $W = \{W_1,\cdots,W_n\}$, where the filter $W_t$ corresponds to the region $M_t$, $t\in[1,n]$.
The feature sizes of the feedback module are listed in Table \ref{tab:feature_size}. Furthermore, we illustrate the generated masks used in Fig. \ref{fig:attention_map} (c). From the figure,  we can observe that the masks can distinguish different semantic regions (e.g., head, background), which is consistent with the HVS.

Based on $M$ and $W$, we perform the region-aware convolution for the stitched image $S$,  which can be formulated as:
\begin{equation}
    F_{u,v} = S_{u,v} \otimes W_t, (u,v)\in M_t,
\end{equation}
where $F_{u,v}$ denote the feature output in the position $(u,v)$; $\otimes$ is the 2D convolution operation. By applying a specific convolution kernel on a specific region, we enable the network to differentiate regions of different semantics, which promotes the extraction of local feature.

\begin{figure}[t]
  \centering

    \includegraphics[width=\linewidth]{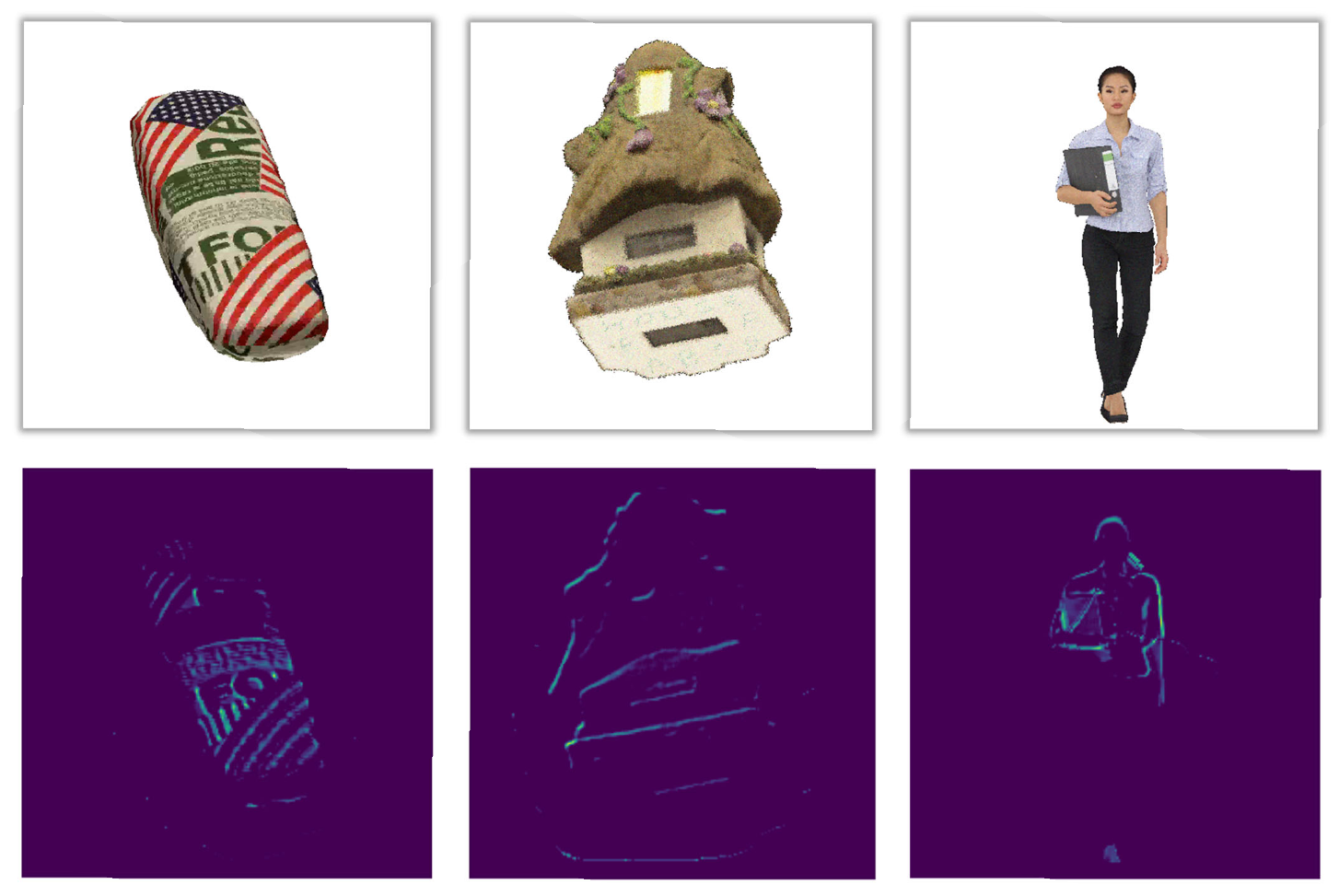}

  \caption{Visualization of local feature maps.}
  \label{fig:local_visualization}
  
\end{figure}

\textbf{Local Feature Generation.} After performing a region-aware convolution for the stitched image $S$, we propose to extract local feature through a shallow convolutional network, seeing Fig. \ref{fig:drconv}. As illustrated in the figure,  we use filters with a very small receptive field ($3\times 3$ and $1\times1$), which makes that the extracted features are localized and usually refer to some image details (e.g., edges, contours). We show some feature maps of the local branch in the Fig. \ref{fig:local_visualization}. The feature maps mainly concern the contour details of point clouds, which is complementary to the semantic information indicated by the attention maps. Finally, we obtain the local feature indicated by $f_l\in\R^{D_o}$.

\subsection{Coarse-to-fine Quality Prediction}

After obtaining the global feature $f_g$ and the local feature $f_l$, to increase the feature discrepancy, we define the feature  disentangling loss as follows:
\begin{equation}\label{eq:loss_cos}
    L_{dis} = \mathrm{max}\{0,\frac{f_g\cdot f_l}{||f_g||_2||f_l||_2}\}, 
\end{equation}
According to Eq. \eqref{eq:loss_cos}, when the cosine similarity between $f_g$ and $f_l$ becomes 0, it indicates that there is no linear correlation between the two features. A margin was given to the loss function because being exactly opposite could harm the representation ability \cite{kim2023feature}.

Then we intend to make a coarse-to-fine quality prediction. Specifically, we define two quality regression heads for two branches:

\begin{equation}
\begin{aligned}
    \hat{q}_{c} &= \mathrm{FC}(f_g), \\
    \hat{q}_{f} &= \mathrm{FC}(f_g\oplus f_l),
\end{aligned}
\end{equation}
where $\mathrm{FC}(\cdot)$ is a full-connection layer; $\hat{q}_c$ and $\hat{q}_{f}$ denote the coarse-grained and fine-grained predicted quality scores respectively. Denoting the subjective mean opinion score (MOS) as $q$, both two predictions are supervised by the regression loss as:
\begin{equation}
\begin{aligned}
    L_{reg} &= \frac{1}{B}\sum_{b=1}^B(\hat{q}_{c}^b-q^b)^2 +\frac{1}{B}\sum_{b=1}^B(\hat{q}_{f}^b-q^b)^2,
\end{aligned}
\end{equation}
where the superscript ``$b$" denotes the $b$-th samples in a mini-batch with the size of $B$. To achieve a progressive quality prediction, we further propose a branch-wise quality ranking loss as:
\begin{equation}\label{eq:loss_rank}
    L_{rank} = \mathrm{max}\{0,\mathrm{SROCC}(\hat{q}_c,q)-\mathrm{SROCC}(\hat{q}_f,q)\},
\end{equation}
where $\mathrm{SROCC}(\cdot)$ denotes the Spearman rank-order correlation coefficient between the predictions and the ground truths in the mini-batch. Note that we refer to \cite{blondel2020fast} to create the differentiable implementation of the ranking metric. Eq. \eqref{eq:loss_rank} enforces $\mathrm{SROCC}(\hat{q}_c,q)<\mathrm{SROCC}(\hat{q}_f,q)$ during training. Such a design can enable networks to take the coarse prediction from the global feature and gradually approach the finer-grained prediction by supplementing the local feature.

Finally,  the overall loss function for training is defined as 

\begin{equation}
    L = L_{reg} + \lambda_1 L_{dis} + \lambda_2 L_{rank},
\end{equation}
where $\lambda_1$ and $\lambda_2$ are the weighting factors.

\section{Experiments}\label{sec:experiment}
\subsection{Databases and Evaluation Protocols}

\textbf{Datasets.} Our experiments are based on three commonly used PCQA datasets, including LS-PCQA Part I\cite{liu2023point}, SJTU-PCQA \cite{yang2020predicting}, and WPC \cite{liu2022perceptual}. LS-PCQA Part I consists of 930 annotated point clouds impaired by 31 types of distortion, which are generated from 85 references. SJTU-PCQA includes 9 reference point clouds and 378 distorted samples impaired with 7 types of distortions (\textit{e.g.,} color noise, downsampling) under 6 levels. WPC contains 20 reference point clouds and 740 distorted samples disturbed by 5 types of distortions (\textit{e.g.,} compression, gaussian noise). 

\textbf{Evaluation Metrics.} Three widely adopted evaluation metrics are employed to quantify the level of agreement between predicted quality scores and the MOS: Spearman rank order correlation coefficient (SROCC), Pearson linear correlation coefficient (PLCC), and root mean square error (RMSE). To ensure consistency between predicted scores and MOSs, a nonlinear Logistic-4 regression is applied to map the predicted scores to the same dynamic range, following the recommendations suggested by the Video Quality Experts Group (VQEG) \cite{Antkowiak2000vqeg}, \textit{i.e.},
\begin{equation}\label{eq:logic_fit}
    \psi(x) = \beta_4+\frac{\beta_1-\beta_4}{1+(x/\beta_3)^{\beta_2}},
\end{equation}
where  $\beta_1$, $\beta_2$, $\beta_3$, and $\beta_4$ are the parameters fitted by minimizing the sum of squared errors.

\begin{table*}[t]
\centering
\caption{Performance Comparison for three database.  The top two performance results are marked in \textbf{boldface} and \underline{underline}.}
\label{tab:overall_performance}
\resizebox{\textwidth}{!}{
\begin{tabular}{c|c|ccc|ccc|ccc}
\toprule
\multirow{2}{*}{Type} &\multirow{2}{*}{Databases} & \multicolumn{3}{c|}{SJTU-PCQA\cite{yang2020predicting}}  & \multicolumn{3}{c|}{WPC\cite{liu2022perceptual}} & \multicolumn{3}{c}{LS-PCQA\cite{liu2023point}}   \\ \cline{3-11}
& & PLCC & SROCC & RMSE & PLCC & SROCC & RMSE & PLCC & SROCC & RMSE  \\ \midrule
\multirow{11}{*}{FR} & MSE-p2po\cite{MPEGSoft}   &0.896	&0.810	&1.046	&0.588	&0.566	&18.385	&0.427	&0.301	&0.744
\\
&MSE-p2pl \cite{MPEGSoft}	&0.783	&0.696	&1.471	&0.482	&0.445	&19.890	&0.454	&0.286	&0.734
\\
&HD-p2po \cite{MPEGSoft} &0.787	&0.708	&1.428	&0.384	&0.258	&21.055	&0.393	&0.273	&0.757
\\
&HD-p2pl\cite{MPEGSoft} &0.755	&0.688	&1.552	&0.371	&0.308	&21.169	&0.405	&0.271	&0.751
\\
&$\rm PSNR_{YUV}$ \cite{MPEGSoft}  &0.764	&0.762	&1.513	&0.574	&0.551	&18.579	&0.527	&0.482	&0.699
\\
&PCQM \cite{meynet2020pcqm} 	&0.836	&0.874	&2.378	&0.702	&0.749	&22.823	&0.208	&0.426	&0.789
\\
&GraphSIM \cite{yang2020inferring}	&0.896	&0.874	&1.040	&0.698	&0.685	&16.171	&0.358	&0.331	&0.767
\\
&MS-GraphSIM \cite{zhang2021ms}	&0.933	&0.903	&0.841	&0.722	&0.713	&15.569	&0.441	&0.404	&0.733
\\
&pointSSIM \cite{alexiou2020pointssim} &0.751	&0.725	&1.539	&0.479	&0.461	&19.905	&0.225	&0.164	&0.804
\\
&MPED \cite{yang2022mped} &0.903	&0.900	&1.000	&0.695	&0.681	&16.313	&0.613	&0.609	&0.646
\\
&TCDM \cite{zhang2023tcdm} &\underline{0.952}	&0.929	&\underline{0.713}	&0.810	&0.806	&13.313	&0.438	&0.413	&0.739
\\ \midrule
\multirow{6}{*}{NR} &PQA-Net\cite{liu2021pqa} &0.898	&0.875	&1.340	&0.789	&0.774	&12.224	&0.607	&0.595	&0.697

\\ 
&ResSCNN\cite{liu2023point} &0.889	&0.880	&0.878	&0.817	&0.799	&13.082	&0.648	&0.620	&0.615

\\ 
&IT-PCQA\cite{yang2022no} &0.609	&0.602	&1.559	&0.581	&0.562	&13.889	&0.447	&0.423	&0.899

\\ 
&GPA-Net\cite{shan2023gpa} &0.901	&0.891	&0.864	&0.796	&0.775	&12.976	&0.623	&0.602	&0.705

\\ 
&MM-PCQA \cite{Zhang2022MMPCQA} &0.939	&0.910	&0.805	&0.846	&0.844	&12.028	&0.644	&0.605	&0.621
\\ 
&GMS-3DQA \cite{zhang2023gms}  &0.916	&0.886	&0.931 &0.818	&0.814	&13.032 &\underline{0.666}	&\underline{0.645}	&\underline{0.606}
\\ 
&3DTA \cite{zhu20243dta}  &0.948	&\textbf{0.931}	&
0.910	&\underline{0.886}	&\underline{0.886}	&\underline{11.636}	&0.609	&0.604	&0.660
\\ 
&\textbf{AFQ-Net}   &\bf{0.957}	&\underline{0.930}	&\bf{0.678}	&\bf{0.897}	&\bf{0.899}	&\bf{9.971}	&\bf{0.690}	&\bf{0.680}	&\bf{0.583}

\\ \bottomrule
\end{tabular}}
\end{table*}

\subsection{Implementation Details}\label{sec:implementation_detail}
We implement our model by PyTorch and conduct
training and testing on the NVIDIA 3090 GPUs. All point clouds are rendered into three types of projected images with a spatial resolution of $512\times512$ by PyTorch3D \cite{ravi2020accelerating}. The used vision transformer is ViT-B \cite{dosovitskiy2020image}.

\textbf{Dataset Split.}  Considering the limited scale of the data set, the k-fold cross-validation strategy is used for the experiment to accurately estimate the performance of the proposed method. Following \cite{Zhang2022MMPCQA}, 9-fold and 5-fold cross-validation is selected for SJTU-PCQA (9 references) and WPC (20 references). For LS-PCQA (85 references), we apply a 5-fold cross-validation.
Note that there is no content overlap between the training and testing sets for the three databases. Specifically, the ratios of content in the training set and test set are 8:1, 16:4, and 68:17 for the three databases. For each fold, the performance on the test set with minimal training loss is recorded, and the average performance across all folds is recorded as the final result.

\textbf{Training Strategy.} All databases are trained for 50 epochs with a batch size of 8. During the training process, all training images are deployed with the traditional data enhancement strategy, \textit{i.e.}, randomly cropping the images to $224\times224$ size such as \cite{su2020blindly,Zhang2022MMPCQA}. During the testing stage, following \cite{su2020blindly}, 10 patches with $224\times 224$ size are randomly sampled and their corresponding prediction scores are average pooled to obtain the final quality score. We use the Adam \cite{kingma2014adam} optimizer with weight decay $5e-4$. The learning rate is set separately as $2e-5$ and $2e-4$ for the pre-trained model and other parts, and is reduced by a rate of 0.9 every 5 epochs. 

\textbf{Network Details.}  To make the projected images compatible with ViT-B, the input patch size of the global branch is $16\times16$. The MSA head number is set by default as $N_h=12$. The number of regions in the guided mask is set to $n=8$ and the kernel size of region-wise filters is set to $k=3$. The global and local feature both have a dimension of $D_o=256$. The weighting factors in the loss function are simply set to $\lambda_1=\lambda_2=1$.

\begin{table*}[pt]
  \centering
  \caption{PERFORMANCE COMPARISON FOR NR-PCQA METRICS ON M-PCCD and BASICS databases}
\resizebox{0.9\textwidth}{!}{
\begin{tabular}{c|cccccccc}

     \hline
    M-PCCD   & PQA-Net & GPA-Net & ResSCNN &IT-PCQA & MM-PCQA & GMS-3DQA &3DTA & AFQ-Net \\
    \hline
    PLCC &0.554	&0.383	&0.298	&0.310
	&\bf{0.829}	&\underline{0.818}	&0.703	&0.816

	\\
    SROCC 	&0.478	&0.396	&0.267	&0.315
	&0.810	&\underline{0.826}	&0.765	&\bf{0.834}

	\\
    RMSE  &0.886	&0.847	&1.004	&0.996	&\bf{0.693}	&\underline{0.702}	&0.995	&0.740
	\\ \hline

\multicolumn{1}{r}{} & \multicolumn{1}{r}{} & \multicolumn{1}{r}{} & \multicolumn{1}{r}{} & \multicolumn{1}{r}{} & \multicolumn{1}{r}{} & \multicolumn{1}{r}{} & \multicolumn{1}{r}{} & \multicolumn{1}{r}{} \\

     \hline
    BASICS   & PQA-Net & GPA-Net & ResSCNN &IT-PCQA & MM-PCQA & GMS-3DQA &3DTA & AFQ-Net \\
    \hline
    PLCC &0.655	&0.391	&0.367	&0.465
	&0.793	&0.895	&\underline{0.896}	&\bf{0.905}

	\\
    SROCC &0.387	&0.352	&0.344	&0.369
	&0.738	&0.807	&\bf{0.825}	&\underline{0.793}

	\\
    RMSE &0.797	&0.975	&0.997	&0.878
	&0.628	&\underline{0.472}	&0.488	&\bf{0.449}
 
	\\

    \hline
    \end{tabular}}
  \label{tab:performance on BASICS}%
\end{table*}

\begin{figure}[t]
  \centering

    \includegraphics[width=\linewidth]{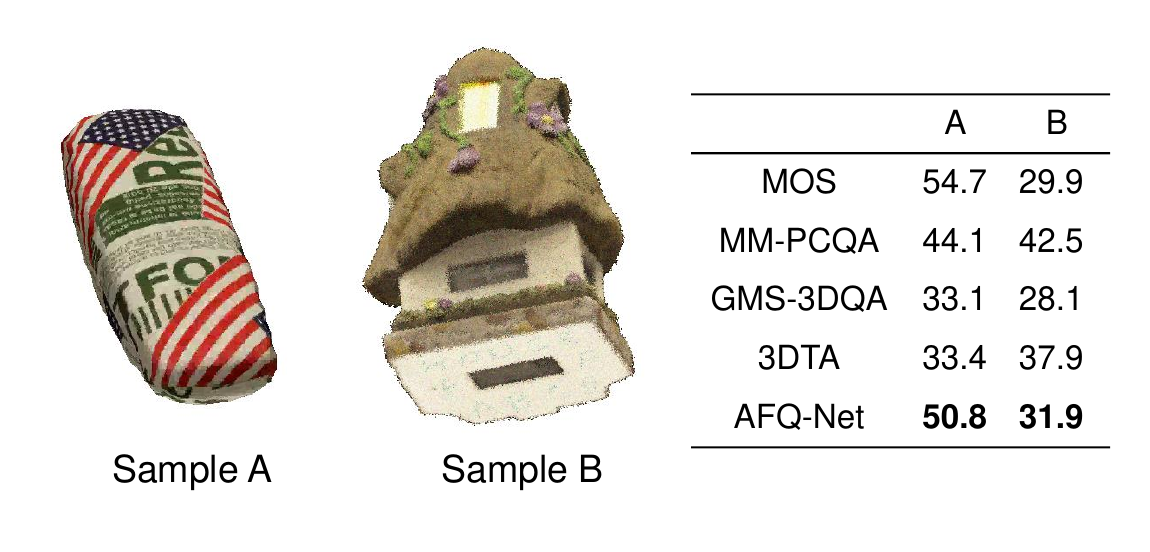}

  \caption{Exemplary point clouds with the subjective MOS and the predicted quality score of different NR-PCQA metrics. The predictions that are closest to the MOS are highlighted in \textbf{boldface}.}
  \label{fig:visual_sample}
  
\end{figure}

\subsection{Performance Comparison}

\textbf{Overall Performance Comparison.}
Table \ref{tab:overall_performance} lists the experimental results on META-3D . The proposed method, AFQ-Net, is compared with 11 FR-PCQA metrics, and 7 NR-PCQA metrics. we run all published codes to produce the results. We strictly retrain the available baselines with the same database split set up to keep the comparison fair. In addition, for the FR-PCQA methods that require no training, we simply validate them on the same testing sets and record the average performance.

\begin{figure*}[t]
  \centering 

    \subfigure[]{
    \includegraphics[width=0.9\linewidth]{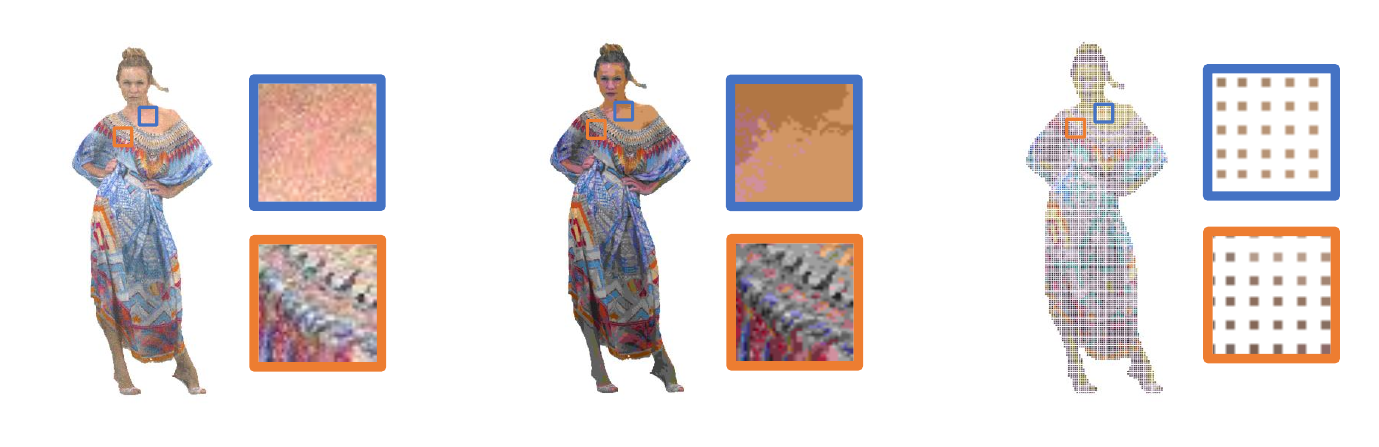}}
    \subfigure[]{
    \includegraphics[width=0.9\linewidth]{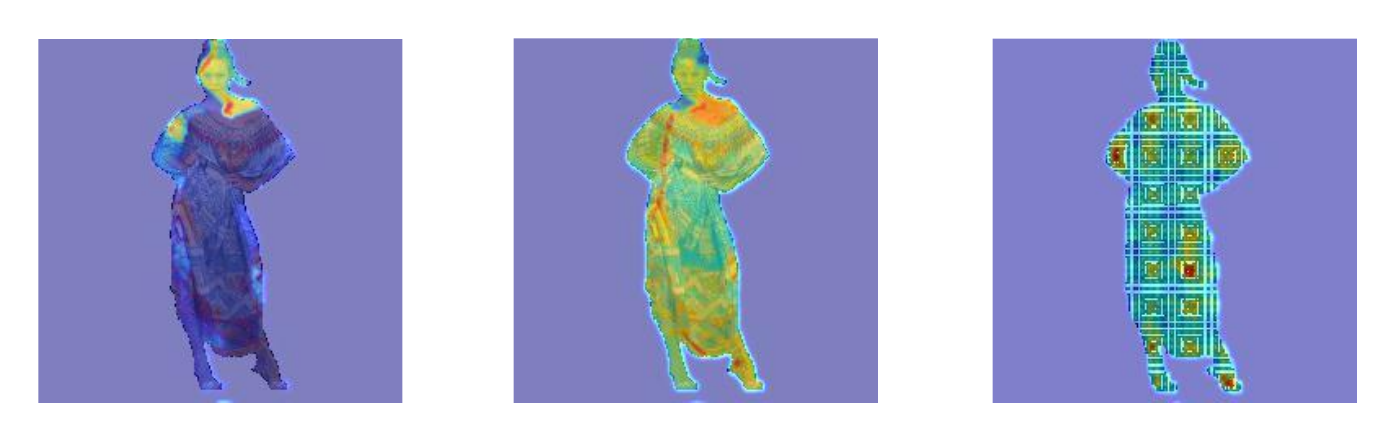}}
  \caption{Visual explanations generated by Grad-CAM++ for three types distortions (Rayleigh noise, V-PCC compression, and G-PCC compression.) (a) Distorted point clouds. (b) Visualization of Grad-CAM++. }
  \label{fig:gradcam}

\end{figure*}

For each database, the top two performance results for each evaluation criterion are highlighted in \textbf{boldface} and \underline{underline}. We can see that the proposed AFQ-Net exhibits outstanding performance across all three datasets. Especially, the proposed AFQ-Net presents remarkable performance on the LS-PCQA, a database with wide diversity in terms of both content and distortion type, which demonstrates the adaptability and robustness of our model.
In comparison, the performance of the existing FR and NR methods varies significantly across different datasets. For example, GMS-3DQA demonstrates strong performance on LS-PCQA but exhibits poor performance on WPC; 3DTA works well on SJTU-PCQA but presents inferior performance on LS-PCQA. IT-PCQA performs poorly because it is based on domain adaptation, in which the labels of point clouds are not utilized in the training. Meanwhile, it should be noted that our method presents competitive performance compared to the FR-PCQA methods despite the inaccessibility of the reference information.

To further validate the model performance in compression scenarios, we test the model performance on two PCQA databases that consist only of compressed samples, \textit{i.e.}, the M-PCCD database (232 distorted samples corresponding to 8 references) \cite{alexiou2019comprehensive} and the BASICS database (1494 distorted samples corresponding to 75 references) \cite{ak2024basics}. 4-fold and 5-fold cross-validations are applied for the two databases, respectively. We report the performance of different NR-PCQA metrics on M-PCCD and BASICS in Tab \ref{tab:performance on BASICS}. According to the table, we can see that AFQ-Net provides the best SROCC on M-PCCD and the best PLCC and RMSE on BASICS, which demonstrates that the proposed method is sensitive to compression distortions. MM-PCQA and 3DTA both presents inferior performance on one database (MM-PCQA on BASICS and 3DTA on M-PCCD). GMS-3DQA works well, which is also based on 2D projection. Based on the results, it seems that metrics using 3D information present unstable performance on different databases. This may be because those metrics only utilize some local patches for quality prediction, which lacks global perception and is easily affected by point density. However, for those metrics that only leverage 2D projection, how to mimic different rendering strategies for different databases has not yet been explored in depth.

We also show some examples of distorted point clouds with the subjective MOS, the predicted score of the proposed AFQ-Net and other NR-PCQA models in Fig. \ref{fig:visual_sample}, where two point clouds are impaired by G-PCC trisoup compression and additive noise. It can be seen that the predicted score provided by AFQ-Net aligns well with subjective perception. In comparison, other models are not sensitive to quality changes because their predicted scores for two point clouds are relatively close. 

\textbf{Qualitative Analysis.}  To demonstrate that our model can attentively
capture the critical features related to distortions, we also present a qualitative analysis using a prevalent tool for network explanation, that is, Grad-CAM++ \cite{chattopadhay2018grad}. As illustrated in Fig. \ref{fig:gradcam}, subfigure (a) presents three samples with the same content but impaired by three different distortions (\textit{i.e.}, Rayleigh noise, V-PCC compression, and G-PCC compression), and subfigure (b) presents their Grad-CAM++ maps that reflect what makes the network perform decision. Note that the extracted activation maps correspond to the last convonlutional layer on the local branch (\textit{i.e.}, the $1\times1$ Conv in Fig. \ref{fig:drconv} (b)). From the figure, we have the following observations: i) the activation map associated with Rayleigh noise distortion mostly highlights these flat regions. This is mainly because Rayleigh noise is located at isolated points. According to texture masking effect \cite{zhang2024perception}, noise in complex texture regions (\textit{e.g.}, orange box) is generally harder to perceive than noise in homogeneous regions (\textit{e.g.}, blue box). ii) For two types of compression distortion, attribute quantization in V-PCC leads to abnormal color mutations, whereas Octree decomposition in G-PCC causes a sparse point distribution. Consequently, the activation maps associated with V-PCC and G-PCC emphasize the whole point cloud because these deformations are severe and not localized. Based on these instances, we
can see that the proposed AFQ-Net can effectively capture and distinguish various distortions.

\begin{table*}[t]
\renewcommand\tabcolsep{7pt}
	\centering

  \caption{Cross-dataset evaluation for NR-PCQA methods. Training and testing are both conducted on complete datasets. Results of PLCC are reported.}
  
	\begin{tabular}{cc|cccccccc}
		\toprule  
		Train & Test & PQA-Net & GPA-Net & ResSCNN &IT-PCQA & MM-PCQA & GMS-3DQA &3DTA & AFQ-Net \\ 
		\midrule  
	LS & SJTU &0.403	&0.618	&0.567 &0.366	&0.715	&\textbf{0.811} &0.671 &\underline{0.777}   \\
	SJTU & LS &0.281	&0.333	&0.338 &0.303	&\bf{0.373}	 &\underline{0.366} &0.288 &0.354 \\
    LS & WPC &0.427	&0.506	&0.435 &0.331	&0.323 &\underline{0.696} &0.545	&\bf{0.753} \\
	WPC & LS &0.285	&0.346	&0.339 &0.278	&\underline{0.515}	 &0.451 &0.386 &\bf{0.625}  \\
    WPC & SJTU &0.255	&0.549	&0.568	&0.395	&\bf{0.740}	&0.517	&\underline{0.726}	&0.558
 \\
	SJTU & WPC &0.266	&\underline{0.420}	&0.271	&0.310	&0.301	&\bf{0.523}	&0.339	&0.385
  \\
		\bottomrule  
	\end{tabular}
	\label{tab:cross}

\end{table*}

\textbf{Cross-Dataset Evaluation.}  The cross-dataset evaluation is conducted to test the generalizability of the NR-PCQA methods when encountering various data distributions. In Table \ref{tab:cross}, we first train the compared models on one database and test the trained model on another database, and then swap the training set and the test set. Note that we use the whole database for training or testing. For each validation, we record the result with minimal training loss and report the results in Table \ref{tab:cross}.
From the table, we can see that: i) AFQ-Net achieves good performance when trained on LS-PCQA, which shows the generalizability of our model when rich content categories are provided. ii) Almost all methods (including AFQ-Net) show poor performance when trained on SJTU-PCQA and tested on LS-PCQA and WPC (denoted by SJTU2LS and SJTU2WPC validation). This is largely due to the limited and extremely biased content in SJTU-PCQA (5 out of 9 contents are about the human body). iii) AFQ-Net trained on WPC shows opposite trends when validated on SJTU-PCQA and LS-PCQA. More specifically, AFQ-Net performs the best when validated on LS-PCQA but presents relatively inferior performance when validated on SJTU-PCQA. This may be due to the large difference in content categories between SJTU-PCQA and WPC (human body \textit{vs.} object). Notably, some recent methods demonstrate superior cross-dataset performance on several validations. For instance, MM-PCQA and 3DTA perform well on the WPC2SJTU validation, likely due to their use of 3D information as input, which enables them to capture common density variations in SJTU-PCQA more effectively. GMS-3DQA achieves the best results on the SJTU2WPC validation, potentially because of its mini-patch sampling strategy. The mini-patch map in GMS-3DQA preserves the distortion patterns of point clouds while blurring content information, which appears advantageous when there are substantial content differences between databases. To enhance the performance of AFQ-Net, a promising approach would be to replace the stitched image in the local branch with the mini-patch map, thus improving its capacity to evaluate local distortions. Additionally, directly applying the asynchronous feedback strategy for 3D data is also a practical and intriguing avenue.

\begin{figure}[t]
  \centering 
    \includegraphics[width=0.9\linewidth]{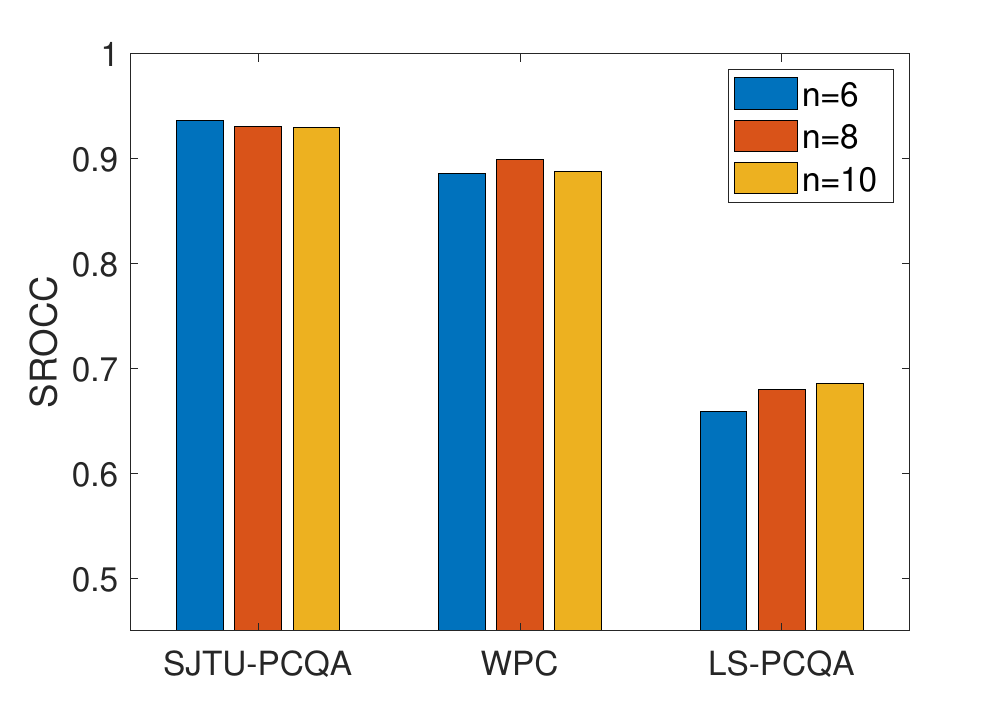}
  \caption{Ablation study for the region count on three databases.}
  \label{fig:barplot_n}

\end{figure}
\begin{table}[t]

  \centering
  \caption{Ablation study for the key modules. `\Checkmark' or `\XSolidBrush' means the setting is preserved or discarded.}

  \resizebox{\linewidth}{!}{
    \begin{tabular}{c|ccc|cc|cc}
    \toprule
    \multirow{2}{*}{Index} & \multirow{2}{*}{Global} & \multirow{2}{*}{Local} & \multirow{2}{*}{Feedback} & \multicolumn{2}{c|}{WPC} & \multicolumn{2}{c}{LS-PCQA}  \\
    &&&& PLCC & SROCC & PLCC & SROCC\\
    \midrule
    (1) & \Checkmark     & \XSolidBrush     & \XSolidBrush  &0.8291	&0.8233	&0.6736	&0.6512

 \\
     (2) & \XSolidBrush    & \Checkmark    & \XSolidBrush  &0.3940	&0.2766	&0.4502	&0.4129
 \\
    (3) & \Checkmark     & \Checkmark     & \XSolidBrush   &0.8701	&0.8692	&0.6786	&0.6662

    \\
    (4) & \Checkmark     & \Checkmark     & \Checkmark    &\bf{0.8974}	&\bf{0.8987}	&\bf{0.6902}	&\bf{0.6796}

  \\
    \bottomrule
    \end{tabular}}%
  \label{tab:ablation_module}%
   \vspace{-0.4cm}
\end{table}%

\subsection{Ablation Studies}

\textbf{Ablation Study for the Key Modules.} To verify the effectiveness of the key modules in the AFQ-Net, we further investigate the contribution of different components and report the results in Table \ref{tab:ablation_module}. In the table, the index (1) and (2) denote using only global branch or local branch for quality prediction respectively. (3) indicates one dual-branch network without feedback connection (\textit{i.e.}, replacing the DRConv in Fig. \ref{fig:drconv} (b) with $3\times3$ convolution), while (4) represents the full model. From the table, we have the following observations: i) Seeing (1) and (2), we see that using only the global branch performs  better than using only the local branch structure. This is mainly because the parameter size of the global
branch is significantly larger than that of the local branch; consequently, the global branch has stronger fitting ability. Meanwhile, the pre-trainning on ImageNet inject the global branch some quality-related prior knowledge, which boosts its performance. ii) Comparing (1) or (2) with (3), the dual-branch structure performs significantly better than the single-branch structure, demonstrating the complementarity of the two features; iii) Seeing (3) and (4), the feedback module further brings a noticeable gain, which verifies the effectiveness of the feedback mechanism in visual quality assessment.

\begin{figure}[t]
  \centering 
    \includegraphics[width=0.9\linewidth]{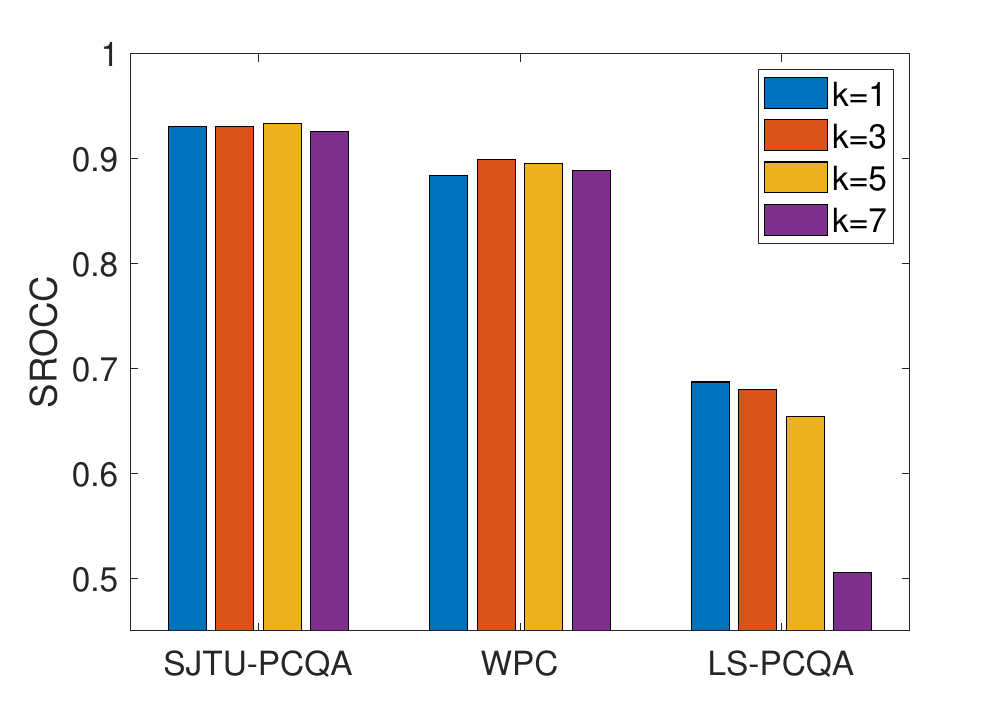}
  \caption{Ablation study for the kernel size on three databases.}
  \label{fig:barplot_k}

\end{figure}

\textbf{Ablation Study for the Region Count.} We test the performance of AFQ-Net under different $n$ values to investigate the influence of the division granularity and report the results in Fig. \ref{fig:barplot_n}. We see that SJTU-PCQA, WPC and LS-PCQA respectively provide the best performance with $n=6,8,$ and $10$. This may be due to the different amount of content in the three databases. SJTU-PCQA has limited and extremely biased content, so few divisions can distinguish different semantic regions while too many divisions cause over-segmentation for the same region. In comparison, WPC and LS-PCQA have more references (20 for WPC and 85 for LS-PCQA) and richer content including different object categories, for which cases large $n$ values can benefit the model performance.

\begin{figure}
    \centering
    \subfigure[]{\includegraphics[width=0.48\linewidth]{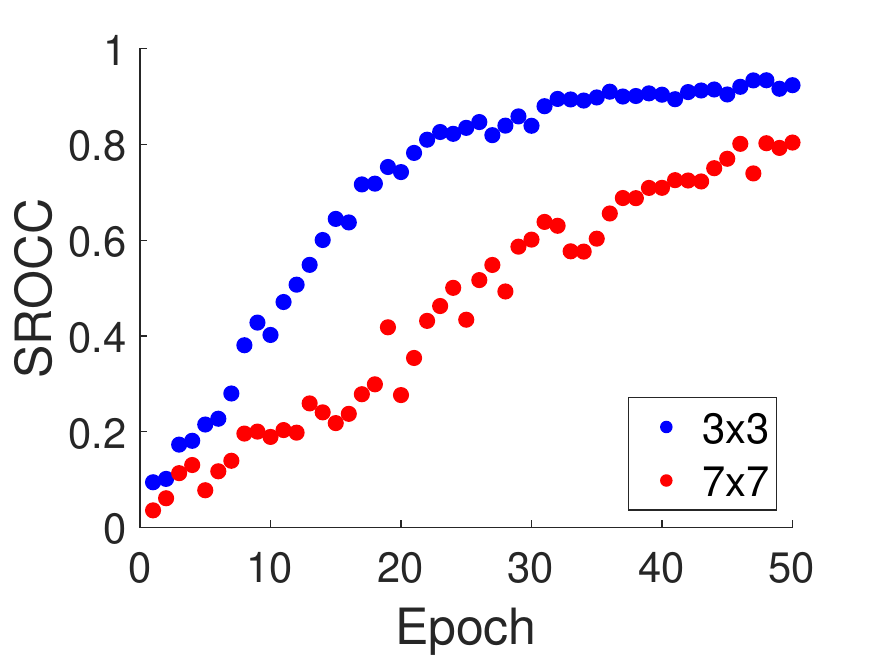}}
    \subfigure[]{\includegraphics[width=0.48\linewidth]{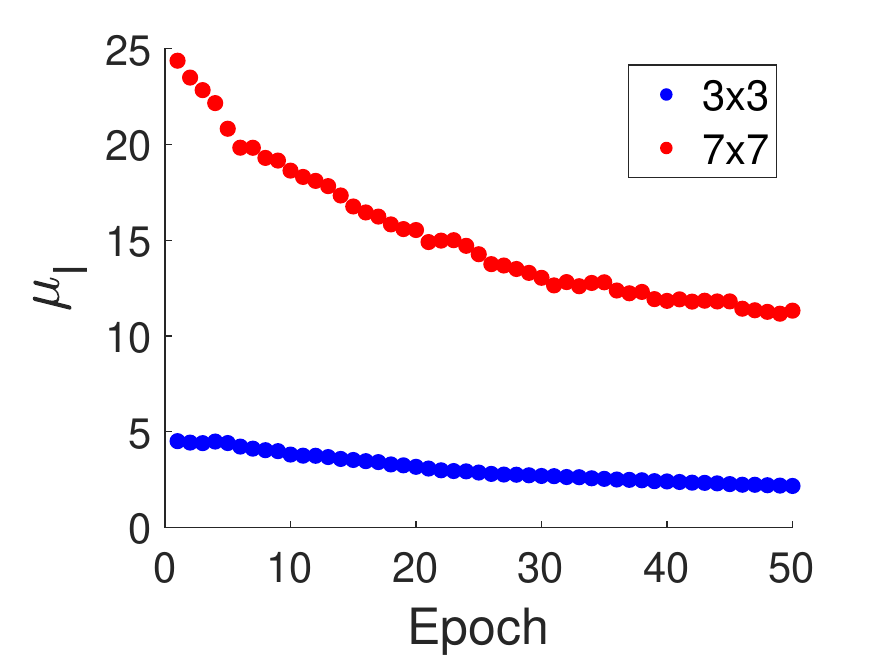}}
    {\includegraphics[width=0.48\linewidth]{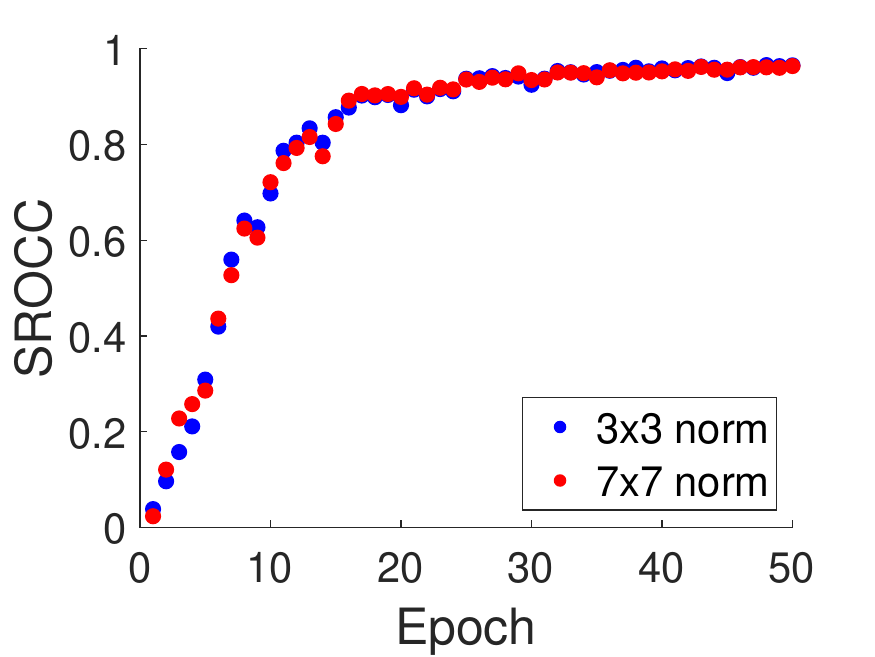}}
    \subfigure[]{\includegraphics[width=0.48\linewidth]{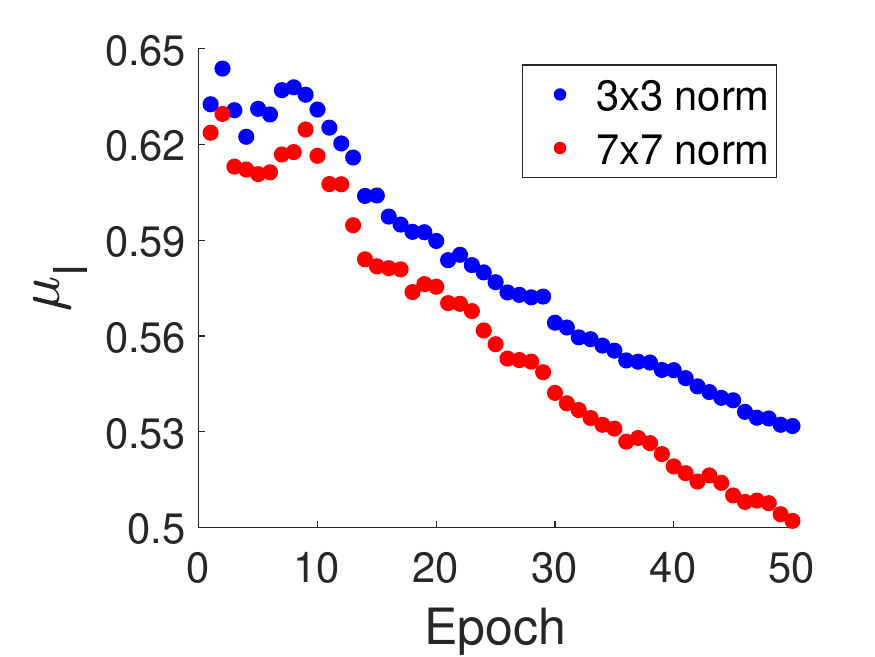}}
    
    \caption{SROCC trend and $\mu_l$ trend versus the training epoch for different kernels. }
    \label{fig:kernel_trend}
\end{figure}

\textbf{Ablation Study for the Kernel Size.} We test the performance of AFQ-Net under different $k$ values to investigate the influence of the kernel size of region-wise filters and report the results in Fig. \ref{fig:barplot_k} (a) and (b). We see that when the kernel size is too large (e.g., $k=7$), the model performance will decrease. Especially, $7\times 7$ kernel causes inferior results on LS-PCQA. Further looking at this case, we find that larger kernel size in DRConv will generate larger element values in the local feature, which will make the local feature contribute more to the prediction, thus affecting the convergence speed of the network. More specifically, given the local feature $f_l$, we define its average absolute value of as $\mu_l=\mathrm{mean}(\mathrm{abs}(f_l))$.We have illustrated the SROCC (on the training set) trend and the $\mu_l$ trend versus the training epoch for $3\times 3$ and $7\times 7$ kernel in Fig. \ref{fig:kernel_trend} (a) and (b). From Fig. \ref{fig:kernel_trend} (b), we can see the $\mu_l$ of $7\times 7$ kernel is significantly larger than that of $3\times 3$ kernel; according to Fig. \ref{fig:kernel_trend} (a), the corresponding SROCC on the training set converges more slowly. In fact, DRConv tends to generate element values in the range of $[0,1]$ for convolution kernels, therefore causing larger results with increasing kernel size. Meanwhile, the complexity of LS-PCQA make the network harder to converge compared to WPC and SJTU-PCQA. A reasonable solution is to add a normalization factor for different kernel size, such as $\frac{1}{9}$ for $3\times 3$ kernel and $\frac{1}{49}$ for $7\times 7$ kernel. We report the SROCC  and the $\mu_l$ trend of two normalized kernels in Fig. \ref{fig:kernel_trend} (c) and (d). We can see the normalized $3\times 3$ and $7\times 7$ kernels present similar trends in both two sub-figures. Meanwhile, the final [PLCC,SROCC,RMSE] criteria for the two normalized kernels are $[0.6826,0.6706,0.5858]$ and $[0.6755,0.6664,0.5962]$ respectively, which demonstrates that the normalization can effectively eliminate the performance drop caused by large kernels.

\textbf{Ablation Study for the Input Image Types.} Three types of image are generated during the preprocessing stage. The depth images serve as a supplement to texture images in the network input stage, while the occupancy images help perform multi-view fusion and region-aware feedback. We conduct experiments to validate the effectiveness of the three types of input and list the results in Table \ref{tab:ablation_input}. In the table, index (1) and (2) represent using only texture or depth images for quality prediction respectively. (3), (4), and (5) denote pairwise combinations of different image types, including ``texture+depth", ``depth+occupancy", and ``texture+occupancy", while (6) indicates the full model. Note that the framework using only occupancy images is infeasible because they are only used to generate some weights and masks. From the table, we can observe: i) based on (1) and (2), model using texture images performs far better than that using depth images. It is reasonable because texture images generally contain visual information than depth images. ii) Comparing (1) with (3), introducing the depth images brings a marginal gain and even impairs the model performance. This is due to the fact that geometry distortions can also be reflected in texture images, therefore reducing the importance of depth information. Especially, most distortion types in LS-PCQA do not cause any geometry deformation, which result in relatively inferior performance when injecting depth information. iii) Seeing (1) and (5), (3) and (6), the introduction of occupancy images leads to noticeable improvements for the three criteria, demonstrating the importance of occupancy information in our model.

\textbf{Ablation Study for the Loss Functions.} The proposed network is trained using the regression loss $L_{reg}$, the feature disentangling loss $L_{dis}$, and the quality ranking loss $L_{rank}$. We evaluate the effect of each loss function and report the results in Table \ref{tab:ablation_loss}. In the table, index (1), (2) and (3) represent using single loss term for quality prediction respectively. (3), (4), and (5) denote pairwise combinations of different loss terms, including $L_{reg}+L_{dis}$, $L_{dis}+L_{rank}$, and $L_{reg}+L_{rank}$, while (6) indicates the full loss. From the table, it is observed: i) seeing (1), (4), and (5), both $L_{dis}$ and $L_{rank}$ benefit the model performance, demonstrating the effectiveness of the two loss functions. ii) According to (6) and (7), the full loss provides noticeable gain over the combination of $L_{reg}+L_{rank}$ on LS-PCQA but presents marginal gain on WPC. The reason may be that $L_{rank}$ also achieves feature disentanglement to some extent, which weakens the effect of $L_{dis}$. Considering $L_{dis}$ only measures the linear similarity between global and local features, one possible avenue for further improvement is to minimize non-linear dependence between two features, such as mutual information.

\begin{table}[t]
  \centering
  \caption{Ablation study for the input image types}
   \resizebox{\linewidth}{!}{
    \begin{tabular}{c|ccc|cc|cc}
    \toprule
    \multirow{2}{*}{Index} & \multirow{2}{*}{Texture} & \multirow{2}{*}{Depth} & \multirow{2}{*}{Occupancy} & \multicolumn{2}{c|}{WPC} & \multicolumn{2}{c}{LS-PCQA}  \\
    &&&& PLCC & SROCC & PLCC & SROCC\\
    \midrule
    (1) & \Checkmark     & \XSolidBrush     & \XSolidBrush   &0.8815	&0.8850	&0.6734	&0.6583
 \\
    (2) & \XSolidBrush      & \Checkmark     & \XSolidBrush   & 0.3374	&0.2128	&0.2354	&0.0485

 \\ \midrule
    (3) &\Checkmark     & \Checkmark     & \XSolidBrush   &0.8821	&0.8826	&0.6655	&0.6508

 \\
    (4) & \XSolidBrush    & \Checkmark      & \Checkmark    &0.3908	&0.2548	&0.2398	&0.0154

    \\
    (5) & \Checkmark     & \XSolidBrush      & \Checkmark &0.8971	&0.8979	&\bf{0.6907}	&0.6744

 \\ \midrule
    (6) &\Checkmark     & \Checkmark     & \Checkmark     &\textbf{0.8974}	&\textbf{0.8987}	&0.6902 &\bf{0.6796}
  \\
  
    \bottomrule
    \end{tabular}}%
  \label{tab:ablation_input}%
 
\end{table}%

\section{Conclusion}\label{sec:conclusion}
In this paper, we aim to solve the problem of global-local feature learning in NR-PCQA and propose an asynchronous feedback network based on the human visual mechanisms. Taking into account the guiding role of global feature in quality perception, our network is implemented by a dual-branch architecture connected by a feedback module. Utilizing the attention maps generated by the global branch, we perform region-aware convolution to derive the local feature. A coarse-to-fine learning strategy is also adopted to fuse the global and local feature into the final quality prediction.  The proposed AFQ-Net shows a consistent and reliable correlation with subjective evaluation on three PCQA benchmarks. Further ablation studies have supported the model design by examining its key modules and parameter values.

\begin{table}[t]
  \centering
  \caption{Ablation study for the loss functions}
   \resizebox{\linewidth}{!}{
    \begin{tabular}{c|ccc|cc|cc}
    \toprule
    \multirow{2}{*}{Index} & \multirow{2}{*}{$L_{reg}$} & \multirow{2}{*}{$L_{dis}$} & \multirow{2}{*}{$L_{rank}$} & \multicolumn{2}{c|}{WPC} & \multicolumn{2}{c}{LS-PCQA}  \\
    &&&& PLCC & SROCC & PLCC & SROCC\\
    \midrule
   (1) & \Checkmark     & \XSolidBrush     & \XSolidBrush   &0.8851	&0.8868	&0.6696	&0.6538

 \\
   (2) &  \XSolidBrush    & \Checkmark    & \XSolidBrush   &0.1985	&0.1133	&0.3097	&0.0236

 \\
    (3) & \XSolidBrush    & \XSolidBrush    & \Checkmark   &0.3724	&0.2317	&0.2428	&0.0660

\\ \midrule
    (4) & \Checkmark     & \Checkmark     & \XSolidBrush    &0.8898	&0.8908	&0.6704	&0.6571

    \\
    (5) & \XSolidBrush     & \Checkmark     & \Checkmark   &0.4247	&0.3418	&0.2802	&0.1379

    \\ 
    (6) & \Checkmark     & \XSolidBrush     & \Checkmark    &0.8970	&0.8981	&0.6704	&0.6600

    \\
    \midrule
    (7) & \Checkmark     & \Checkmark     & \Checkmark     &\textbf{0.8974}	&\textbf{0.8987}	&\bf{0.6902}	&\bf{0.6796}\\
    
    \bottomrule
    \end{tabular}}%
  \label{tab:ablation_loss}%

\end{table}%

\ifCLASSOPTIONcaptionsoff
  \newpage
\fi

\bibliographystyle{IEEEtran}
\bibliography{ref}

\end{document}